\documentclass[11pt, a4paper, logo]{deepmind}

\usepackage[authoryear, sort&compress, round]{natbib}
\title{Selection-Inference: Exploiting Large Language Models for Interpretable Logical Reasoning}

\correspondingauthor{tonicreswell@deepmind.com}

\usepackage{hyperref}       
\usepackage{url}            
\usepackage{booktabs}       
\usepackage{amsfonts}       
\usepackage{nicefrac}       
\usepackage{algorithm}
\usepackage{algpseudocode}
\usepackage{microtype}      
\usepackage{graphicx}
\usepackage{subcaption}
\usepackage{color}
\usepackage{amsmath}
\usepackage{amssymb}
\usepackage{multirow}
\usepackage{wrapfig}
\usepackage{xcolor}
\usepackage{wrapfig}

\newcommand{\totaltaskn}{50 }
\newcommand{\sitaskn}{10 }

\usepackage{listings}
\newcommand{\beginsupplement}{%
        \setcounter{table}{0}
        \renewcommand{\thetable}{S\arabic{table}}%
        \setcounter{figure}{0}
        \renewcommand{\thefigure}{S\arabic{figure}}%
     }

\author[1]{Antonia Creswell}
\author[1]{Murray Shanahan}
\author[1]{Irina Higgins}

\affil[1]{DeepMind}

\begin{abstract}
Large language models (LLMs) have been shown to be capable of impressive few-shot generalisation to new tasks. However, they still tend to perform poorly on multi-step logical reasoning problems. Here we carry out a comprehensive evaluation of LLMs on \totaltaskn tasks that probe different aspects of logical reasoning. We show that language models tend to perform fairly well at single step inference or entailment tasks, but struggle to chain together multiple reasoning steps to solve more complex problems. In light of this, we propose a Selection-Inference (SI) framework that exploits pre-trained LLMs as general processing modules, and alternates between selection and inference to generate a series of interpretable, casual reasoning steps leading to the final answer. We show that a 7B parameter LLM used within the SI framework in a 5-shot generalisation setting, with no fine-tuning, yields a performance improvement of over 100\% compared to an equivalent vanilla baseline on a suite of \sitaskn logical reasoning tasks. The same model in the same setting even outperforms a significantly larger 280B parameter baseline on the same suite of tasks. Moreover, answers produced by the SI framework are accompanied by a \emph{causal} natural-language-based reasoning trace, which has important implications for the safety and trustworthiness of the system.
\end{abstract}

\begin{document}

\maketitle

\section{Introduction}
Large language models (LLMs) are powerful few-shot learners \citep{bommasani2021opportunities, brown2020language, lu2021pretrained}. However, one area where they tend to perform poorly is logical reasoning \citep{rae2021scaling}. Yet the ability to perform multi-step, logically valid reasoning is fundamental for the discovery of new knowledge and explainability. It underpins many advancements that have been made in science, medicine, maths and philosophy. It is also one of the most valued strengths of classical, symbolic AI over contemporary deep learning methods \citep{marcus2019rebooting, marcus2020next,bengio2021deep}, prompting the recent increase in the use of neurosymbolic approaches to bridge this gap \citep{garnelo2019reconciling,garcez2020neurosymbolic}. Here we propose a Selection-Inference (SI) framework that takes inspiration from the neurosymbolic literature to improve the ability of LLMs to do logically valid reasoning.

There are many flavours of neurosymbolic models \citep{garcez2020neurosymbolic}. Those from which we draw inspiration tend to have a modular structure, where each module is specialised for one type of operation \citep{mao2019neuro, andreas2016neural}. For example, such modules may be neural networks or hand-crafted functions designed to attend to a single object, or to compare the location or size of two inputs \citep{andreas2016neural, yi2018neural}. Neurosymbolic models can produce an answer to a complex query by chaining these operations together, passing inputs from one module to another. This has the benefit of producing an interpretable trace of intermediate computations, in contrast to the ``black-box'' computations common to end-to-end deep learning approaches. Importantly, the modularity of neurosymbolic methods allows them to generalise to significantly harder problems that require long chains of reasoning \citep{hudson2019learning}. However, the hand-crafted and specialised nature of the modules often makes the resulting systems brittle, hard to optimise, and difficult to extend to new domains \citep{yi2018neural}.

\begin{figure}[t!]
    \centering
    \includegraphics[width=0.85\textwidth]{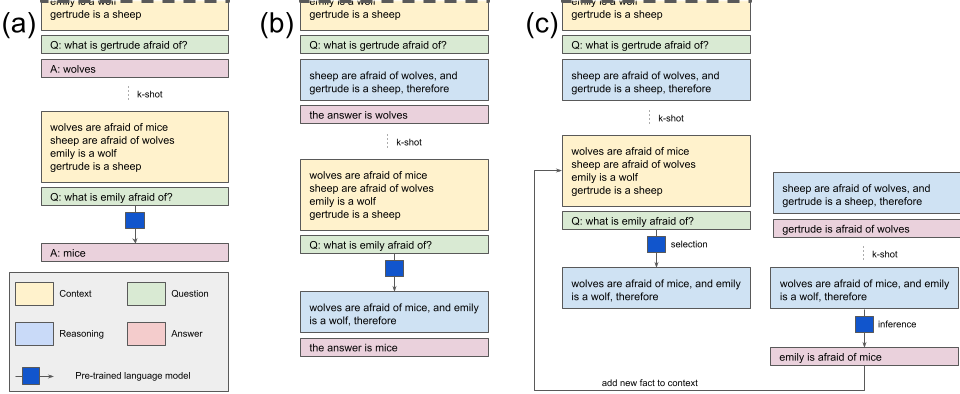}
    \caption{Selection-Inference (SI) framework (c) in comparison with the vanilla baseline (a) and Chain-of-Thought, COT, (b) approach  \citep{wei2022chain}. \textbf{(a)}: The vanilla large language model baseline takes in concatenated [\emph{context}, \emph{question}, \emph{answer}]$\times k$ for k-shot prompting, followed by [\emph{context}, \emph{question}] and is asked to generate the \emph{answer}. All reasoning is done implicitly; \textbf{(b)}: COT \citep{wei2022chain} inspired baseline takes in concatenated [\emph{context}, \emph{question}, \emph{reasoning}, \emph{answer}]$\times k$ for k-shot prompting, followed by [\emph{context}, \emph{question}] and is asked to generate the [\emph{reasoning}, \emph{answer}]; \textbf{(c)}: SI framework consists of two steps. The selection step takes in concatenated [\emph{context}, \emph{question}, \emph{selection}]$\times k$ for k-shot prompting, followed by [\emph{context}, \emph{question}] and is asked to select a subset of facts from the context to support a single step of \emph{reasoning}. The inference step takes in [\emph{selection}, \emph{inference}]$\times k$ for k-shot prompting, followed by the \emph{selection} produced by the Selection module to produce a new fact (the \emph{inference}) to be added to the context. Each combination of [selection + inference + add fact to context] makes up one step of reasoning. These can be chained together to answer harder problems. The final \emph{inference} is taken to be the \emph{answer}.}
    \label{fig:SI_framework}
\end{figure}

Our SI framework, drawing on best practice from neurosymbolic approaches, decomposes logical reasoning into two modular stages: 1) \emph{selection}, which involves choosing a subset of relevant information sufficient to make a single step of inference, and 2) \emph{inference}, which only sees the limited information provided by the selection module, and uses it to infer a new intermediate piece of evidence on the way to the final answer (see Fig.~\ref{fig:SI_framework}c). We implement both stages using pre-trained LLMs which, thanks to their powerful few-shot generalisation capabilities, serve as more general alternatives to the hand-crafted, specialised modules typically used in neurosymbolic approaches. In the SI framework, multiple steps of selection and inference are chained together to produce a sequence of reasoning steps. As well as underpinning better performance on reasoning problems, this yields an interpretable trace that justifies the final answer.

Furthermore, the reasoning trace produced by our system is \emph{causal}, in the sense that each step follows from, and depends on, the previous step. Each inference step is made in isolation, based solely on the limited information provided by the Selection module, without direct access to the question or to previous steps of reasoning. This contrasts with the more common approach of obtaining \emph{post-hoc rationalisation}, where the answer produced by the model has no direct dependence on the explanation, since the explanation is produced either in parallel to the answer or after the fact \citep{saha2020prover, lampinen2021tell, cobbe2021training}. A notable example that sits in the grey area between post-hoc rationalisation approaches and the more causal explanation approaches is Chain-Of-Thought (COT) \citep{wei2022chain} (see Fig.~\ref{fig:SI_framework}b). In this approach LLMs are encouraged to produce a reasoning trace before the answer. However the dependence of the answer on the reasoning is not explicitly encouraged to be causal (as defined above). Indeed, the authors show that while the COT explanations help boost the final answer accuracy, the reasoning traces produced by the model are often wrong even when the final answer is correct (see the Supplementary Materials of \citep{wei2022chain} for examples). Developing a system that can demonstrate how it reaches its answers using a \emph{causal} reasoning trace has important benefits in terms of safety, explainability, interpretability, debugging, and trust. In this paper we make the following contributions:
\begin{enumerate}
    \item We provide a comprehensive evaluation of LLMs on a set of \totaltaskn tasks probing different aspects of logical reasoning, and show that LLMs are good at simpler single step logical inference in 5-shot generalisation settings, but struggle with harder problems (Sec.~\ref{sec:how_well_do_llms_reason})
    \item We introduce the Selection-Inference (SI) framework, a modular, iterative approach to solving reasoning problems (Sec.~\ref{sec:selection_inference}).
    \item We demonstrate the utility of the SI framework by evaluating a 7B parameter LLM from the Gopher family \citep{rae2021scaling} on \sitaskn logical reasoning tasks, showing overall that it almost triples the performance of the same model used naively and almost doubles the performance of the same model used in the COT framework. Moreover, it often outperforms a 40x larger 280B LLM baseline used both naively and in the COT framework. 
    \item We illustrate further benefits of the SI framework in terms of the causal and interpretable reasoning traces produced (Sec.~\ref{sec:si_reason_traces}). These traces can help humans understand how the model reached its final answer, which is useful for debugging and opens the system's decisions to human critique.
\end{enumerate}

\begin{figure}
\begin{subfigure}[b]{0.49\textwidth}
    \centering
    \includegraphics[width=0.8\textwidth]{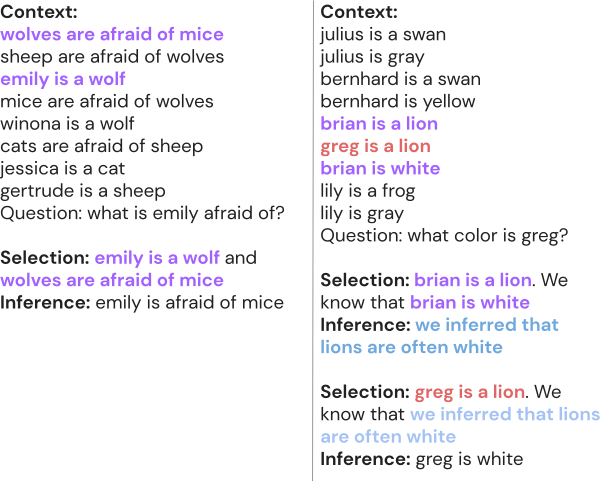}
    \caption{Correct reasoning on bAbI deduction (left) and induction (right) tasks.}
    \label{fig:si_example_babi} 
    \end{subfigure}
    \hspace{0.5cm}
\begin{subfigure}[b]{0.49\textwidth}
    \includegraphics[width=0.99\textwidth]{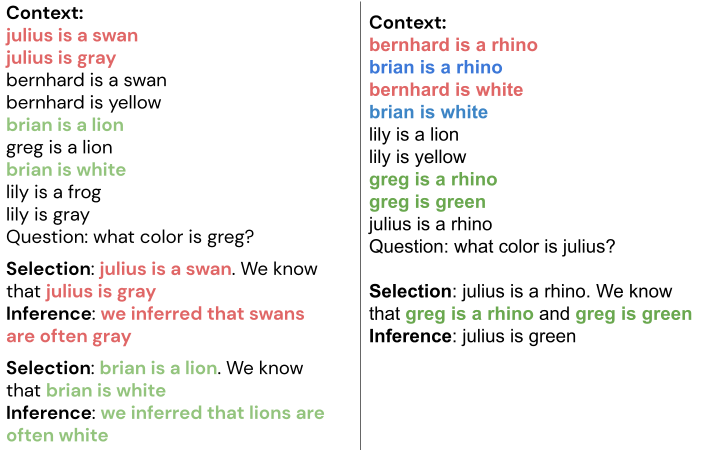}
    \caption{SI can recover from an error (left) and justify an ambiguous answer with a reasoning trace (right).}
    \label{fig:recover_from_error}
\end{subfigure}
\caption{Qualitative results from the Selection-Inference (SI) model on bAbI tasks.}
\label{fig:si_qualitative_results}
\end{figure}

\section{Related Work}
Our Selection-Inference framework sits at the intersection of classical, symbolic AI and deep learning. A typical symbolic AI system might consist of a knowledge base, which is typically hand-curated by experts, and an inference engine that allows the system to perform logic-based reasoning over its knowledge base. For example, such a system could apply step-by-step reasoning to answer a complex question such as ``What are the apothecary's incentives and disincentives for selling poison to Romeo in Romeo and Juliet?'' \citep{lenat_2019} - something that even the best contemporary deep learning based systems struggle to do.

One of the primary benefits of symbolic AI systems over deep learning models is their interpretability; we can look at the reasoning steps such a system has taken to see how the final conclusion was reached. However, unlike deep learning approaches, symbolic AI systems require knowledge to be hand-crafted and are in general hard to scale. Although some approaches have attempted to bridge the gap between deep learning and symbolic AI by converting problems into formal logic \citep{nye2021improving} and using existing solvers to help produce an answer, this process can be brittle and tends not to scale well. Another attempt to bridge the gap comes from the neurosymbolic perspective \citep{mao2019neuro, yi2018neural, gupta2019nmn, hudson2019learning}. These models combine the best parts of deep learning -- learning knowledge from data -- and symbolic AI -- producing an interpretable reasoning trace. However, they are typically quite brittle due to the hand-crafted \citep{gupta2019nmn}, specialised nature \citep{mao2019neuro} of the modules and optimisation difficulties.

On the deep learning side, recent work has attempted to adapt large pre-trained language models, LLMs, to the task of logical reasoning. At a high level these can be split into three groups: 1) approaches that try to fine-tune LLMs to produce the final answer directly, keeping reasoning implicit \citep{betz2020critical, clark2020transformers} (e.g. Fig.~\ref{fig:SI_framework}a); 2) approaches that encourage LLMs to produce reasoning explicitly, but all reasoning steps are produced in one generative step \citep{nye2021show, zelikman2022star,jhamtani2020learning,Dalvi2021ExplainingAW,wei2022chain,cobbe2021training} (e.g. Fig.~\ref{fig:SI_framework}B); and 3) approaches that use LLMs to produce each reasoning step one at a time \citep{tafjord2020proofwriter}. The latter is where our Selection-Inference framework sits (Fig.~\ref{fig:SI_framework}C). In general it was found that the approaches that incorporate explicit reasoning work better than those that only try to predict the final answer. However, although explicit reasoning helps improve the accuracy of the models, encouraging the models to produce multiple steps of reasoning in a single generative pass is not enough to make the models use reasoning in a causal manner. The authors found that the generated reasoning traces often contain unrelated or incorrect steps while still resulting in the correct answer (see examples in the appendices of \citep{zelikman2022star, wei2022chain}). Encouraging LLMs to generate each reasoning step one at a time \citep{tafjord2020proofwriter} is currently the most promising direction for achieving causal reasoning, and it is the approach we take in our paper. While the model proposed by \cite{tafjord2020proofwriter} is very impressive, it only works for ``Prove this statement to be True/False'' style questions, since it relies on enumerating all possible implications and checking whether the question statement or its negation are present in the inferred facts, which is also computationally expensive. 

\section{How Well Do Large Language Models Reason?}\label{sec:how_well_do_llms_reason}

Past work has shown that LLMs are poor at logical reasoning \citep{rae2021scaling}, however the evaluation was done on a relatively small set of tasks, and was not systematic. In particular, here we are interested in 1) how LLMs perform on simple entailment tasks compared to multi-step reasoning problems and 2) how scaling laws -- suggested by \cite{rae2021scaling} -- apply to logical reasoning. To this end, we evaluated LLMs from the Gopher family in a 5-shot\footnote{We chose 5-shot setting because it was used in \citep{rae2021scaling}, and because \citep{min2022rethinking} have demonstrated that additional shots beyond 4 result in limited increase in multi-choice accuracy} setting on a larger set of \totaltaskn tasks that touch on different aspects of logical reasoning and vary in terms of the number of reasoning steps required, presence or absence of negation, whether the relevant context information was provided, and whether the model is required to evaluate the accuracy of multiple choices or generate the answer among others. The additional tasks were collected from six sources: bAbI \citep{weston2015towards}, BigBench \citep{BigBench}, AAC \citep{betz2020critical}, Jeopardy \citep{tunguz_2019}, ProofWriter \citep{tafjord2020proofwriter} and 2WikiMultiHop \citep{welbl2018constructing} (see Fig.~\ref{fig:logic_baselines} in Supplementary Information for raw results).

\begin{figure}[t]
    \centering
    \subfloat[Scaling laws for natural language tasks (bigbench, dark purple line, squares, 56 tasks) and tasks involving logical reasoning (logic, light purple line, circles, 38 tasks). All accuracy results are calculated relative to the random baseline (0\% accuracy means chance level). Only multi-choice tasks are used. \label{fig:scaling_laws}]{
        \includegraphics[width=0.46\textwidth]{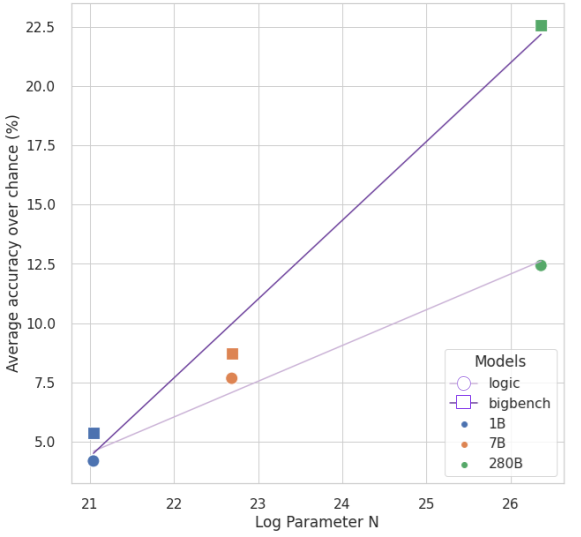}
    }
    \hspace{0.5cm}
    \subfloat[Language models perform well for simple entailment tasks (AAC tasks, Entailed Polarity), their performance starts to get worse on single step inference problems (bAbI task 1, ProofWriter tasks 0-1), and they struggle with more complex multi-step reasoning problems (2WikiMultiHop tasks, bAbI tasks 2-3, ProofWriter tasks 2-5, StrategyQA).\label{fig:baseline_multistep_reasoning}]{
        \includegraphics[width=0.46\textwidth]{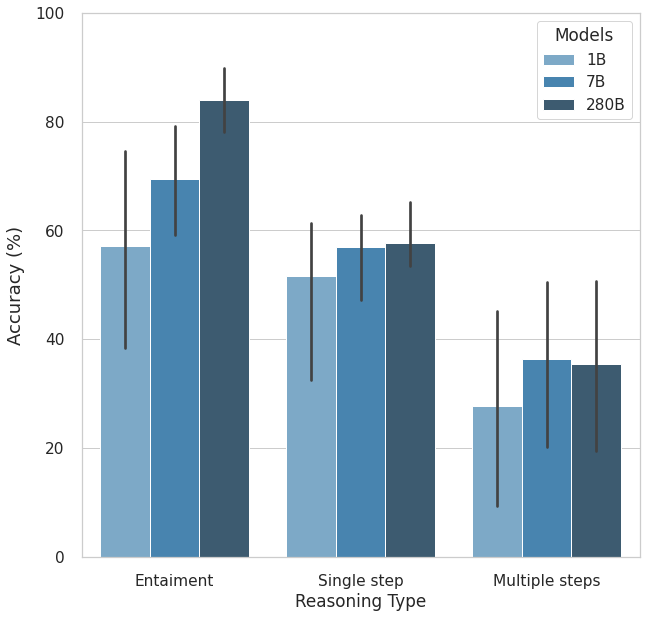}
        }
    \caption{Vanilla language models perform poorly on multi-step logical reasoning tasks.}
    \label{fig:baseline_results}
\end{figure}

Our analysis found that LLMs are good at some aspects of logical reasoning (Fig.~\ref{fig:baseline_multistep_reasoning}). In particular, they appear to be good at simple entailment and implication tasks (e.g. see AAC tasks and Entailed Polarity in Fig.~\ref{fig:logic_baselines}). This appears to hold when negation is present (AAC Split Extended tasks), and both in generative (AAC Split) and multiple-choice scoring settings (AAC Split Extended tasks, Entailed Polarity). However, the performance of vanilla language models tends to decrease when they get presented with irrelevant facts alongside the ones relevant for reasoning (e.g. see 2WikiMultiHop With Context tasks, bAbI tasks 2-3 or ProofWriter tasks), when they have to infer the relevant facts from memory (e.g. 2WikiMultiHop or StrategyQA tasks), and as the questions start to require more steps of reasoning (e.g. see the performance drop between bAbI tasks 1-3 or ProofWriter Tasks).

In line with Rae et al.'s findings, our results confirmed that LLMs of larger sizes do perform better than the smaller models. However, we found that even the largest 280B model performed only 13.6\% above chance level on average across the 38 available multi-choice logic tasks  (see Figs.~\ref{fig:logic_baselines}-\ref{fig:logic_baselines_multichoice_less_random} and Sec.~\ref{si_sec:baseline_results} in Supplementary Information for more details).
Furthermore, we found that logical reasoning tasks were qualitatively different from other natural language tasks. The scaling law for logic-based tasks within the Gopher family models was significantly worse than for other language tasks measured here as the average performance on the subset of BigBench tasks from \citep{rae2021scaling} with the logic tasks used in this paper removed (see Fig.~\ref{fig:scaling_laws}). 

\section{The Selection-Inference (SI) Framework} \label{sec:selection_inference}

We are interested in solving logical reasoning problems expressed in natural language. In this work we assume that each question is accompanied by context information (see Figs.~\ref{fig:SI_framework} and \ref{fig:si_example_babi}), which contains all the information necessary to solve the problem, as well as potentially irrelevant distractors. In the future this assumption can be relaxed, for example by extracting the necessary information through search \citep{lazaridou2022internet, menick2022teaching}. We also assume that all questions are well posed and definitively answerable given the context. 

Logical reasoning problems require using existing information to infer new relevant knowledge necessary to answer the question. This can be done through deduction, induction or abduction, although the datasets we use here contain mostly deductive and a small number of inductive problems\footnote{See Fig.~\ref{fig:si_qualitative_results} for an example of deduction and induction problems used in this paper.}. Some problems require multiple steps of inference, where later steps use the knowledge inferred in the earlier steps. Hence, we use an iterative framework where at each step the SI uses information in the existing context, $\mathcal{C}_t$, to infer a new fact, $f_t$, which is appended back to the context to create new context, $\mathcal{C}_{t+1} = \mathcal{C}_t \cup f_t$. This process can then iterate until the solution to the question is found. In the current implementation of the SI framework, we repeat the process for a fixed number of steps and take the final inference to be the answer. Addressing the issue of halting is left for future work.

Inspired by neurosymbolic methods, we additionally split each step of reasoning into further two components: 1) \emph{Selection}, which selects a subset of the information present in the context, $s_t$, given the context and the question, $\mathcal{C}^t \cup q$. This selection, $s^t$ is fed to the next step, 2) \emph{inference}, which produces the new fact, $f_t$, based on the information passed to it by the selection step. Examples of selection and inference are shown in Fig.~\ref{fig:si_example_babi}. This splitting of each step of reasoning into selection and inference is the main contribution of our paper, and is important for several reasons. First, and most importantly, it makes the resulting reasoning \emph{causal}, since both steps have limited capabilities by design, and are interdependent. The selection step is constrained to only use the information available in the context, an the inference step only sees the subset of facts provided by the selection without access to the question. Hence, the model is unlikely to make up information to answer the question, and it cannot ignore reasoning when producing the final answer. The other benefit of this approach is that each step of reasoning is broken down into even smaller sub-tasks, which are easier for LLMs to adapt to, and which helps make the reasoning more generalisable to harder problems. 

In this paper we use pre-trained, frozen language models from the Gopher family \citep{rae2021scaling} in a 5-shot generalisation setting using prompt engineering to implement the Selection and Inference modules. We settled on prompt engineering to evaluate the base utility of the SI framework, however it can also be used in the fine-tuning setting which we explore briefly in Sec.~\ref{sec:fine_tuning}. We next describe the Selection and Inference modules in more detail.

\subsection{Selection Module}\label{sec:selection_module}

We use prompt engineering to encourage the model to output the correct selection, $s^t$. The n-shot prompt is a string of the the following form:

\begin{small}\begin{lstlisting}
"""
# n-shot prompt
# First example.
<context 1>
<question 1>
# Example selection
<fact>. We know that <fact>[ and <fact>]*. Therefore,
...
# Problem to solve.
<context>
<question>

"""
\end{lstlisting}\end{small}
where \verb|<fact>|s are copied directly from the \verb|context|, and \verb|[ and <fact>]*| means that the module is allowed to select more than one fact for each step of inference, where the total number of facts is a hyper-parameter. 

The simplest option to implement the selection is to feed this prompt directly to a pre-trained LLM and take the output generated by the language model. However, this unconstrained approach may allow the model to make up facts, thus removing the causal aspect of the reasoning trace. Indeed during experimentation this is what we often found. So instead we use the pre-trained LLM to score each of the facts in the context, and select the one with the highest log-likelihood. We can then repeat this process by appending each new fact to the end of the previous prompt until the full selection is constructed. Note that for now we \texttt{halt} after a fixed number of steps. See Algorithm \ref{alg:scoring_selection} for more details.

\subsection{Inference module} \label{sec:inference}

The n-shot prompt for the Inference module has the following form (shown below, also see Fig.~\ref{fig:SI_framework}):

\begin{small}\begin{lstlisting}
    """
    #n-shot inference prompt
    # First example.
    <fact>. We know that <fact>[ and <fact>]*. Therefore, <new fact>.
    ...
    
    # Problem to solve.
    <output of the Selection module>. Therefore,
    """
\end{lstlisting}\end{small}

The n-shot prompt and the output of the Selection module, are fed to the pre-trained LLM serving as the Inference module. The first generated sentence (extracted from the generated text as per BigBench evaluation \citep{BigBench} pipeline, see Supplementary Materials for details) is taken to be the newly inferred fact. This fact is added to the context, which concludes one reasoning step of the SI framework. For now, we halt after a fixed number of steps. See Algorithm \ref{alg:selection_inference} for more details.

\begin{small}\begin{algorithm}
\caption{Selection-Inference}\label{alg:selection_inference}
\begin{algorithmic}
\Require \texttt{An n-shot selection prompt,} $p_{select}$.
\Require \texttt{An n-shot inference prompt,} $p_{infer}$.
\Require \texttt{Initial Context,} $\mathcal{C}^0$, \texttt{made up of facts (and rules).}
\Require \texttt{The question,} $q$.
\Require \texttt{Language model, LLM.}
\Require \texttt{A halting function, halt, determines if the answer has been reached.}
\State $t=0$ \Comment{Start at step $0$}.
\While{\texttt{not halt()}}
    \State $s^t \gets$ \texttt{Selection\_Module}$(p_{select}, C^t, q, \texttt{LLM})$ \Comment{Do selection.}
    \State $i^t \gets$ \texttt{Inference\_Module}$(p_{infer}, s^t)$ \Comment{Do inference.}
    \State $C^{t+1} \gets C^t \cup i^t$ \Comment{Add the newly inferred fact to the context.}
    \State $t \gets t+1$ \Comment{Move onto the next step of reasoning}

\EndWhile \\
\Return{$s^t$}
\end{algorithmic}
\end{algorithm}
\end{small}

\section{Experiments and Results}
\label{sec:experiments_results}
We evaluate our SI framework on a subset of \sitaskn/\totaltaskn logical reasoning tasks introduced in Sec.~\ref{sec:how_well_do_llms_reason}. These tasks were chosen based on whether they include context information necessary to answer the question, whether the questions have a definitive answer, and to ensure that they cover different kinds of reasoning abilities. The tasks include bAbI \citep{weston2015towards} Tasks 1-3, which require the model to use 1-3 supporting time-ordered facts respectively to answer a question, and Tasks 15-16, which test deductive and inductive reasoning respectively. We also evaluate our model on the ProofWriter OWA datasets \citep{tafjord2020proofwriter} of depth 0, 1, 2, 3 and 5 (there is no depth 4 task). These are language-based logical reasoning problems, where the depth is the number of reasoning steps required to answer the question.

To baseline the performance of the SI framework, we consider a 7B (same size as the LLM used in the SI framework) and a 40x larger 280B parameter LLM evaluated in a 5-shot setting. There are two types of evaluation for these vanilla baselines that we consider: \textit{multi-choice} and \textit{generative} evaluation. In \textit{generative} evaluation, we measure the exact string match (first sentence in lower case and ignoring any non-alphabetic characters) between the output generated by the LLM and the ground truth answer. This is appropriate, since most of the tasks that we consider require either a single word answer, or the dataset is such that the answers are highly structured. In \textit{multi-choice} evaluation the LLM is used to score each of the answer choices, as in Li et al. \citep{li2021systematic}. In general LLMs perform significantly better in a \textit{multi-choice} vs \textit{generative} evaluation setting, since the chance level in the multi-choice setting is significantly higher.

We also consider a chain-of-thoughts (COT) \citep{wei2022chain} inspired baseline, where the k-shot prompts to the 7B and 280B models include reasoning traces for the same examples that we use to prompt the SI framework (although with selection and inference combined, see Supplementary Information for example prompts). This tests whether providing the reasoning examples alone is sufficient to improve performance, or whether the further breakdown into Selection and Inference sub-steps improves accuracy. Note that among all of the approaches outlined only the SI framework is explicitly set up to generate \emph{causal} reasoning traces. 

\begin{figure}[t]
    \centering
    \subfloat[\label{fig:summary_results} Average accuracy over 11 datasets comparing like-for-like generative performance of the 7B and 280B parameter language models used in a 5-shot generalisation setting to predict the answer directly (vanilla), in the Chain-Of-Thought framework, COT, and in the SI framework. ]{
        \includegraphics[width=0.35\textwidth]{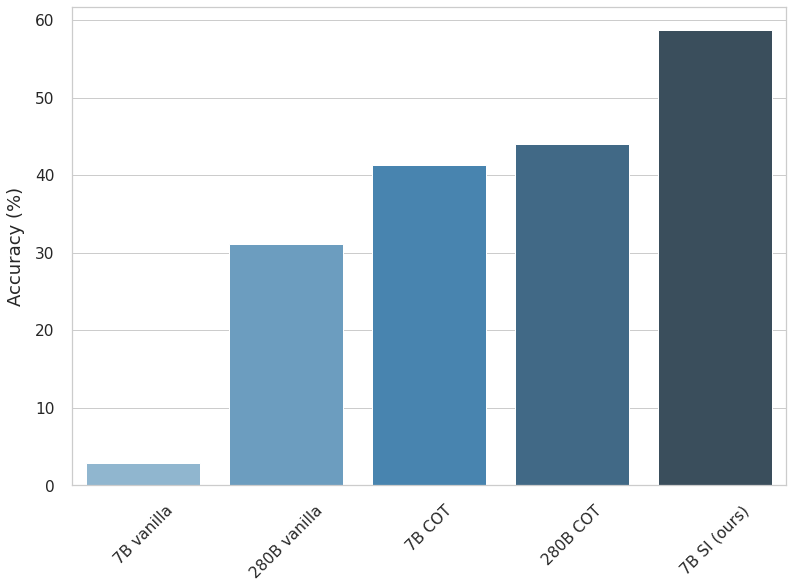}
        }
    \hspace{0.5cm}
    \subfloat[\centering \label{fig:quant_results} Per task breakdown of the performance of LLMs used naively, and within the COT and SI frameworks.]{
        \includegraphics[width=0.55\textwidth]{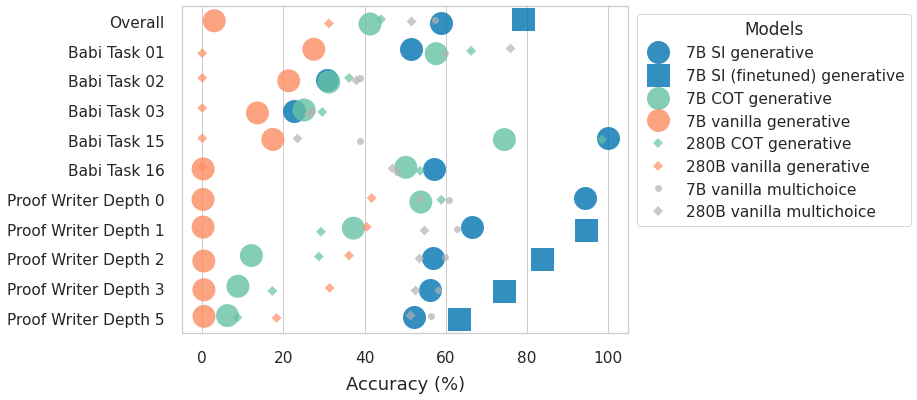}
    }        
    \caption{Quantitative results for the Selection-Inference (SI) framework.}
    \label{fig:si_results}
\end{figure}

Fig.~\ref{fig:summary_results} demonstrates that overall when evaluated generatively, the 7B parameter LLM within the SI framework performs better (58.75\%) than the same model evaluated naively (2.94\%) or in the COT framework (41.32\%) (all $p<0.01$, see Supplementary Information for details of the calculations). Not only that, the 7B parameter LLM within the SI framework also outperforms on average the 40x larger 280B parameter LLM in both vanilla (31.19\%) and COT framework (44.03\%) (all $p<0.01$). When evaluated in the easier multi-choice setting, we find that surprisingly\footnote{This \emph{could} suggest that the 280B LLM has stronger priors, than the 7B LLM, which it favours over logical reasoning. For example, favouring \texttt{sheep are afraid of wolves} despite a context to the contrary \citep{min2022rethinking}. However this requires further investigation.} the vanilla 7B parameter LLM outperforms the 280B parameter LLM (57.31\% vs 51.45\%), while still performing significantly worse than the 7B SI model ($p=0.012$). Note that the latter is evaluated in the harder generative setting. Per task breakdown shown in Fig.~\ref{fig:quant_results} demonstrates that the SI framework solves the bAbI 15 deduction task, the only model to achieve 100\% accuracy (significant difference from the other models, $p<0.01$). Furthermore, it does so having seen only five examples in the prompt. The 7B SI model also significantly outperforms all other models on ProofWriter Depth 0 ($p<0.01$), ProofWriter Depth 1 ($p=0.034$).

As well as improving upon most baselines quantitatively, the SI framework also has additional qualitative benefits: 1) it produces a causal, human interpretable reasoning trace that shows how the model reached its answer and 2) it is able to recover from errors. We will now discuss each of these in turn. 

Since the Selection module is only allowed to pick facts from the context and is separate from the Inference module, and since the latter does not have access to the question, the model has to use what is selected and cannot bypass the selection to compute its answer, thus creating a causal reasoning trace. Since the reasoning trace is in natural language and is causal, it can be audited and inspected by humans, which has significant implications for safety and interpretability. 

Example reasoning traces produced by the SI model are shown in Fig.~\ref{fig:si_example_babi}. In the bAbI 16 example shown on the right the model is solving an inference problem, which requires the model to infer the colour of an animal given facts about the colours of other animals. In this example, the model is asked \texttt{"What colour is greg"}, and told that \texttt{"greg is a lion"}. This means first the model needs to use induction to infer a rule about the colour of lions. On the first step, we see that the model induces a rule, \texttt{"lions are white"}, based on the fact that \texttt{"brian is a lion"} and \texttt{"brian is white"}; we can see exactly what data underlies the model's decision to form a new rule. On the second step, we see that the model applies this newly inferred rule to the fact that \texttt{"greg is a lion"} to reach the final conclusion that \texttt{"greg is white"} using deduction. Note that the ability of the SI framework to produce inductions relies on its ability to deal with uncertainty and understand what is ``reasonable'' - something that LLMs are naturally capable of, while also being a known struggle point for symbolic AI. 

Since the reasoning traces are output in natural language, they are easy for humans to interpret and potentially intervene. Consider a scenario where there may be both a white lion and a green lion mentioned in the context, in which case we could see which information the model used to make its final decision and decide whether we want to trust it (example in Fig.~\ref{fig:recover_from_error}). We could also imagine examples where the model puts together two unrelated facts to come up with an incorrect inference, and this could also be easily be spotted by a human and rectified by replacing the wrongly inferred fact with a correct one, and re-running the consequent reasoning steps. 

Aside from inspecting reasoning traces and using them to debug when something goes wrong, the additive nature of our model - it accumulates new knowledge with each reasoning step, means that it also has the ability to recover from errors. Fig.~\ref{fig:recover_from_error} demonstrates such an example. In the first step the model inferred that \texttt{"swans are often gray"}, using the facts that \texttt{"julius is a swan"} and \texttt{"julius is gray"}. While this is correct, this new information is not useful for answering the question, which asks about lions. However, it is still possible for the model to make the more useful inference that \texttt{"lions are often white"} in a later step and recover from its original misstep.

\section{Fine-tuning Language Models for Selection and Inference} \label{sec:fine_tuning}
In Sec.~\ref{sec:experiments_results} we have demonstrated significant improvements in logical reasoning accuracy when using prompt-engineering to specialise LLMs for Selection and Inference in the SI framework. Prompt-engineering has the additional benefit of not requiring large amounts of step-by-step reasoning data, which may be hard to obtain. 
In this section we investigate whether the accuracy of the SI framework can be further improved by fine-tuning the LLMs for Selection and Inference.
We use the ProofWriter dataset for which ground truth reasoning traces exist. 

The Selection LLM is fine-tuned to select a subset of sentences (including facts and rules) from the context by generating a string of the form \emph{"sent 2. We know that sent 4 [and sent 7]*."} given the context and the question. We ask the Selection LLM to generate sentence labels (e.g. \emph{"sent 2"} or \emph{"sent 4"}) instead of the sentences themselves, because this prevents the Selection LLM from cheating by making up facts to answer the question quicker. This preserves the dependency of the selection and therefore subsequent reasoning steps on the context. The Inference LLM is fine-tuned to compute an entailment given the selection. Both models are fine-tuned on single steps of reasoning only. Example training data are shown in Fig.~\ref{fig:fine_tune_train_data}. 

\begin{figure}
    \centering
    \begin{subfigure}[l]{0.45\textwidth}
         \centering
         \includegraphics[width=\textwidth]{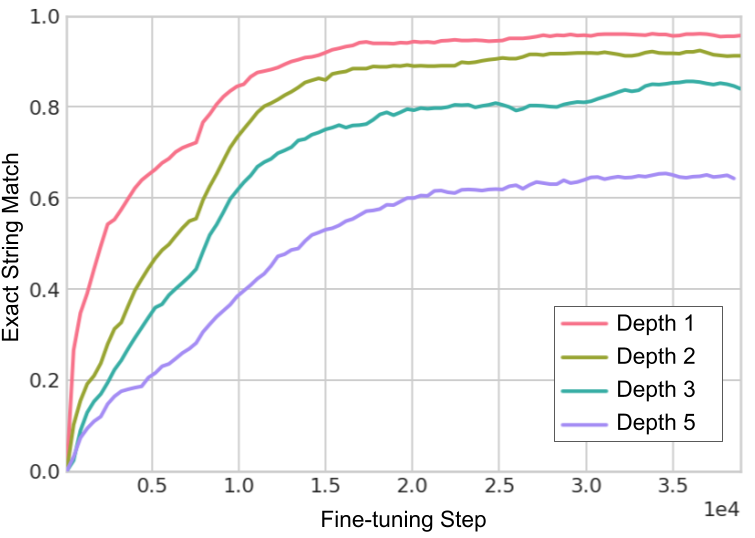}
         \caption{Average test fine-tuning accuracy for the Selection module trained on single-step selection across all ProofWriter datasets (depth 1, 2, 3 and 5) and tested on problems of each depth separately.}
         \label{fig:fine_tune_select}
    \end{subfigure}
    \hspace{0.5cm}
    \begin{subfigure}[c]{0.45\textwidth}
    \centering
    \includegraphics[width=0.7\textwidth]{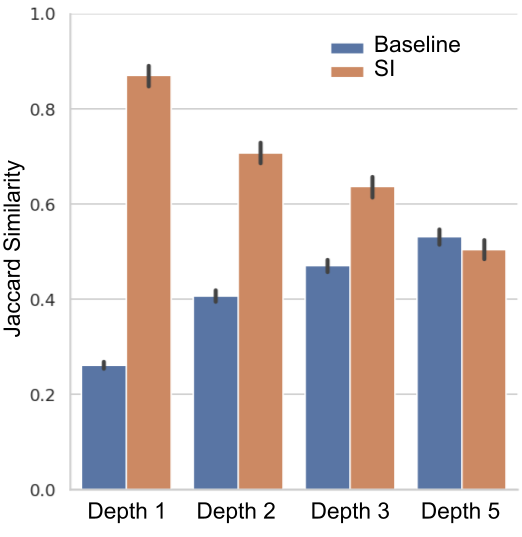}
    \caption{Intersection over union between reasoning traces produced by a model and the ground truth reasoning steps. Baseline, 7B parameter LLM fine-tuned to predict all reasoning steps in one go; SI, using same 7B LLM fine-tuned for single step Selection and Inference.}
    \label{fig:fine_tune_vs_baseline}
    \end{subfigure}
    \caption{Fine-tuning the SI framework on the ProofWriter dataset.}
\end{figure}

The Inference LLM converged very quickly to >99\% test accuracy after only 300 fine-tuning steps with a batch size of 16, which is to be expected as we found that pre-trained LLMs are good at single step entailment out of the box as shown in Fig.~\ref{fig:baseline_multistep_reasoning}. Examples of inferences made by the model are shown in Fig.~\ref{fig:fine_tune_inf}. The Selection LLM was trained for $4\times10^4$ steps (with batch size 16 for 50 hours on a TPU) with the exact string match accuracy reported in Fig.~\ref{fig:fine_tune_select}. Although we notice that the model is much better at predicting selections for problems that require fewer steps of inference than those that require more, ultimately the model still achieves high ($>80\%$) accuracy across most of the reasoning depths. Predicting early selections for deeper reasoning problems is hard, because it requires planning. It is an important problem to address in future work.

Fig.~\ref{fig:quant_results} shows that fine-tuning LLMs on single steps of reasoning within the SI framework leads to significant improvements in final reasoning accuracy (78.95\%) over the prompt-engineered version of the SI framework (57.93\%) and other prompt-engineered baselines (vanilla/COT generative 7B: 0.34/15.73\%, 280B: 31.58/21.12\%). We also found that the fine-tuned 7B LLM used within the SI framework produces significantly more accurate reasoning traces compared to the same LLM fine-tuned to predict all reasoning steps in one go (Fig.~\ref{fig:fine_tune_vs_baseline}). We quantified this using the Jaccard Similarity, $\texttt{Jaccard Similarity} = (M \cap GT) / (M \cup  GT)$, between the proof steps predicted by each model, $M$, and the ground-truth reasoning steps, $GT$, as shown in, calculated using exact string match over alphanumeric characters.

Qualitatively we observed that while the baseline model is good at predicting most of the reasoning steps, they often appear in the wrong order, there are additional reasoning steps that are not on the minimal reasoning path, and some steps get repeated a number of times. 

\section{Conclusion}\label{sec:conclusion}
We have presented the Selection-Inference framework for improving the ability of pre-trained language models to solve logical reasoning problems expressed in natural language. Our approach borrows from the best practices of neurosymbolic approaches to break down logical reasoning into a modular recursive pipeline that not only significantly improves the reasoning accuracy, but also produces a \emph{causal} interpretable reasoning trace. We have demonstrated that prompt-engineered LLMs used in the SI framework significantly outperform both the vanilla and COT baselines evaluated in equivalent settings, and even 40x larger baselines. The performance of the SI framework can be further improved through fine-tuning if step-by-step reasoning data is available.

A model capable of casual, interpretable and logically valid multi-step reasoning has potential applications in law, medicine, science, maths, economics, and other areas where trustworthy and verifiable logical inference is important. At the same time we recognise that special care will need to be taken to evaluate such a model before deployment in any of these settings. Further work is also needed, for example, to improve the Selection module (e.g. by allowing the model search over and evaluate different reasoning traces); to address the halting issue (both in terms of when to stop the selection and when to stop the overall reasoning); to incorporate verifiers that would help avoid false inferences being added to the context; to enable the system to source its own relevant context rather than relying on it being provided in the dataset; and to extend the ability of the system to deal with ambiguous or unanswerable questions.

\bibliography{main}

\clearpage

\appendix
\beginsupplement
\section*{Supplementary Information}
\section{Example prompts for vanilla baselines}

\subsection{ProofWriter}
\begin{lstlisting}
"""
Here are some statements that describe a situation:
Bob is cold.
Charlie is quiet.
Gary is cold.
Harry is quiet.
Big things are cold.
All blue things are not cold.
If something is quiet and blue then it is not cold.
All quiet things are cold.
If something is big and rough then it is round.
If something is cold and not rough then it is blue.
If something is quiet and not furry then it is not blue.
Round things are big.
Based on the above, the statement "Charlie is cold" is true.

...

Here are some statements that describe a situation:
Erin is not cold.
Erin is kind.
Erin is red.
Erin is smart.
Erin is not white.
Erin is young.
Gary is cold.
Gary is not furry.
Gary is kind.
Gary is red.
Gary is not smart.
Gary is young.
All cold, smart things are not furry.
Young, cold things are not furry.
If something is white and smart then it is furry.
If Gary is white then Gary is not furry.
If Erin is young then Erin is furry.
If Gary is not young then Gary is smart.
If Erin is cold then Erin is young.
Red things are kind.
Based on the above, the statement "Erin is not furry" is
"""
\end{lstlisting}

\subsection{bAbI 1}
\begin{lstlisting}
"""
Context: daniel went to the bedroom
daniel journeyed to the office
daniel travelled to the bathroom
mary went to the office
john journeyed to the bedroom
daniel went back to the kitchen
john went to the garden
daniel travelled to the office
Question: where is john?
Choice: garden
Choice: bathroom
Choice: office
Choice: kitchen
Choice: bedroom
Choice: hallway
Answer: garden

...

Context: sandra went to the kitchen
sandra went to the office
sandra travelled to the hallway
sandra went back to the kitchen
mary travelled to the hallway
sandra went to the bedroom
john went to the garden
sandra travelled to the office
Question: where is sandra?
Choice: garden
Choice: bedroom
Choice: kitchen
Choice: bathroom
Choice: hallway
Choice: office
Answer:
"""
\end{lstlisting}

\subsection{2WikiMultiHop}

New lines are added between facts to fit on the page.

\begin{lstlisting}
"""
    Q: When did Michael Baden-Powell's father die?
Here are some relationships to help answer this question.
Michael Baden-Powell::father::Peter Baden-Powell, 2nd Baron Baden-Powell,
Peter Baden-Powell, 2nd Baron Baden-Powell::date of death::9 December 1962
A: 9 December 1962

...

Q: Where does Ekaterina Rybolovleva's father work at?
Here are some relationships to help answer this question.
Ekaterina Dmitrievna Rybolovleva::father::Dmitry Rybolovlev,
Dmitry Rybolovlev::employer::Uralkali
A:
"""
\end{lstlisting}

\section{Example prompts for COT baselines}
\subsection{ProofWriter 3}
\begin{lstlisting}
"""
Given a set of rules and facts, you have to reason whether a statement is true or false.
Here are some facts and rules:
If someone is red then they are nice.
If someone is kind and red then they are white.
If someone is nice then they are kind.
Fiona is red.
Does it imply that the statement "Fiona is not white" is True?
Reasoning: If someone is red then they are nice. We know that Fiona is red. Therefore, Fiona is nice.
If someone is nice then they are kind. We know that Fiona is nice. Therefore, Fiona is kind.
If someone is kind and red then they are white. We know that Fiona is kind and Fiona is red. Therefore, Fiona is white.

...

Here are some facts and rules:
If someone chases the cow then they eat the cow.
If someone is big then they chase the cow.
If someone needs the bald eagle then the bald eagle is big.
If the bear is nice and the bear needs the cow then the bear eats the lion.
If someone needs the lion and they eat the bald eagle then they are blue.
If someone eats the bear and they do not chase the cow then the cow is young.
the bald eagle eats the lion.
the bear is round.
the lion eats the bald eagle.
the bald eagle needs the cow.
the bear is young.
the cow is not nice.
the cow does not chase the bald eagle.
the bear does not eat the bald eagle.
the bear needs the bald eagle.
the bald eagle chases the bear.
Does it imply that the statement "The bald eagle does not eat the cow" is True?
Reasoning: If someone needs the bald eagle then the bald eagle is big. We know that the bear needs the bald eagle. Therefore, the bald eagle is big.
If someone is big then they chase the cow. We know that the bald eagle is big. Therefore, the bald eagle chases the cow.
If someone chases the cow then they eat the cow. We know that the bald eagle chases the cow. Therefore, the bald eagle eats the cow.
"""
\end{lstlisting}

\subsection{bAbI 2}
\begin{lstlisting}
"""
Below are some stories about people moving objects between rooms. After each story you have to answer a question about where a particular object is.
Story:
at t=0 mary grabbed the football there 
at t=1 daniel got the apple there 
at t=2 mary went to the kitchen 
at t=3 daniel journeyed to the office 
at t=4 daniel went to the bedroom 
at t=5 mary moved to the garden 
Question: where is the apple?
Reason: at t=1 daniel got the apple there. We know that at t=4 daniel went to the bedroom. Therefore, the apple is in the bedroom

...

Story:
at t=0 sandra went to the office
at t=1 john took the milk there
at t=2 sandra got the milk there
at t=3 john dropped the milk
Question: where is the milk?
Reason: at t=2 sandra got the milk there. We know that at t=0 sandra went to the office. Therefore, the milk is in the office
"""
\end{lstlisting}

\section{Example prompts for Selection-Inference}\label{sec:example_prompts}

\subsection{bAbI 2}\label{sec:babi2_prompt}

The \textbf{selection} prompt:

\begin{lstlisting}
"""
Here are a collection of stories about people carrying objects from one room to another. You will be asked where any object is. To answer this question you need to figure out who last had the object and which room they have the object in by the end of the story. Here are some examples:

Story:
at t=0 mary grabbed the football there 
at t=1 daniel got the apple there 
at t=2 mary went to the kitchen 
at t=3 daniel journeyed to the office 
at t=4 daniel went to the bedroom 
at t=5 mary moved to the garden 
Question: where is the apple?
Reason: at t=1 daniel got the apple there. We know that at t=4 daniel went to the bedroom

...

at t=0 john moved to the bathroom
at t=1 john travelled to the office
at t=2 john picked up the football there
at t=3 john journeyed to the bathroom
Question: where is the football?
Reason:"""
\end{lstlisting}

The \textbf{inference} prompt:

\begin{lstlisting}

"""
at t=1 daniel got the apple there. We know that at t=4 daniel went to the bedroom. Therefore, the apple is in the bedroom.

...

at t=2 john picked up the football there. We know at t=0 john moved to the bathroom. Therefore"""
\end{lstlisting}

\subsection{ProofWriter}\label{sec:proof_writer_si_prompt}

Below is an example \textbf{selection} prompt. Note that this is for a depth-2 problem and so we show examples of the first reasoning step where the conclusion would not directly prove or disprove the statement and the last reasoning step, where the conclusion would directly prove or disprove the statement.

\begin{lstlisting}
"""
Given a set of rules and facts, you have to reason whether a statement is true or false.

Here are some facts and rules:
Nice people are quiet.
If Dave is smart then Dave is nice.
All white people are smart.
Dave is smart.
Harry is cold.
Does it imply that the statement "Dave is not quiet" is true?
Reasoning: If Dave is smart then Dave is nice. We know that Dave is smart. Therefore,


Here are some facts and rules:
Blue things are green.
All blue things are white.
If Anne is not big then Anne is blue.
Big things are white.
All kind things are round.
If something is white and big then it is not kind.
If something is big and not rough then it is green.
If something is white and blue then it is not green.
Erin is not white.
Anne is big.
Bob is rough.
Anne is white
Does it imply that the statement "Anne is kind" is True?
Reasoning: If something is white and big then it is not kind. We know that Anne is white and Anne is big. Therefore,

...

Here are some facts and rules:
If something likes the squirrel and it is not young then it chases the lion.
If something likes the squirrel then it is rough.
If something chases the rabbit and the rabbit is not young then it chases the lion.
If something eats the lion then it is young.
If something likes the rabbit then it chases the rabbit.
All rough things are nice.
the rabbit is young.
the squirrel likes the rabbit.
the lion likes the squirrel.
Does it imply that the statement "The lion is not nice" is True?
Reasoning:"""
\end{lstlisting}

Example \textbf{inference} prompt:

\begin{lstlisting}
"""
Nice people are quiet. We know that Dave is nice. Therefore, Dave is quiet.

...

If the cow chases the bald eagle then the cow eats the bald eagle. We know that the cow chases the bald eagle. Therefore"""
\end{lstlisting}

\section{Selection-Inference evaluation details}\label{sec:si_eval_details}

\subsection{Selection module}

The algorithm for the Scoring Selection module is shown in Algorithm \ref{alg:scoring_selection}.
\begin{small}\begin{algorithm}
\caption{Scoring \texttt{Selection\_Module }} \label{alg:scoring_selection}
\begin{algorithmic}
\Require \texttt{An n-shot prompt,} $p$.
\Require \texttt{Initial Context,} $\mathcal{C}^0$, \texttt{made up of facts (and rules).}
\Require \texttt{The question,} $q$.
\Require \texttt{Language model, LLM.}
\Require \texttt{A halting function, halt, determines if the selection is complete.}
\State $s^t \gets \texttt{empty string}$
\While{\texttt{not halt()}}

    \State
    
    \State $s_{\texttt{temp}} \gets \arg \max_{\texttt{rule\_or\_fact} \in \mathcal{C}} \sum_{\texttt{token}\in \texttt{rule\_or\_fact} } \texttt{LLM}(\texttt{token}| p, \mathcal{C}^t, q, s^t)$ \Comment{Choose the rule or fact with the maximum log-likelihood under the LLM model.}
    
    \State

    \State $s^t \gets \texttt{join}(s^t, s_{\texttt{temp}})$ \Comment{Join the selected fact or rule to the selection string.}
    
\EndWhile
\Return{$s^t$}
\end{algorithmic}
\end{algorithm}\end{small}

\subsection{Inference module}
To extract the new fact to be added to the context we filter out the first sentence of the text generated by the LLM using the following regular expression: $\texttt{r`[\^{}.?!; }\backslash \texttt{n]+'}$.

\subsection{bAbI}
For all bAbI tasks, the answer is a single word. For example, in bAbI 1-3 the answer is one of \texttt{["hallway", "bathroom", "bedroom", "garden", "kitchen", "office"]}; for bAbI 15 the answer is one of \texttt{["sheep", "cat", "mouse", "wolf"]} and for bAbI 16 the answer is one of \texttt{["yellow", "gray", "green", "white"]}. However, our model outputs a complete sentence, for example \texttt{"emily is afraid of mice"}. Therefore, we take the answer to be the final word output by the inference model on its last step. 

To obtain the results for bAbI tasks 1-3, 15 and 16 shown in Fig.~\ref{fig:quant_results} we prompted the language model to solve the problem in a single step of reasoning. An example of such a prompt is shown in Sec.~\ref{sec:babi2_prompt}. 

We run the SI model for only a single step of reasoning too. Additional steps may increase the chance of the model reaching the correct answer, however, we do not yet have a mechanism for halting reasoning when the answer is reached.

BAbI 16 is an inductive reasoning task that \emph{could} be solved in two steps (rather than one). The first step requires a rule to be inferred and the second step requires the inferred rule to be applied to another fact. For this reason, we also apply SI for two steps to solve the bAbI 16 problems, first inferring a rule from a number of facts and then applying the rule to the correct fact. An example of this is shown in Fig.~\ref{fig:si_qualitative_results}. Using this two step approach, we can see exactly which facts contributed to the formation of a new rule.

\subsection{ProofWriter}
We use a subset of the ProofWriter Open World Assumption, OWA,  dataset. In the Close World Assumption dataset, CWA, everything that can be proven is True otherwise it is False. This means things are either True or False. This also means that reasoning traces are only provided when a statement is True, but not when a statement is False. To "show" something is False one has to enumerate all possible facts (possibly up to a certain depth) and then if a statement is not shown to be True it is assumed to be False. It is therefore not simple to generate meaningful reasoning traces for these types of problems.

On the other hand, in the OWA data if it is not possible to prove something is True or False, then it is Unknown. This means that for True and False examples, where one may want to show \texttt{p(x)}, reasoning traces are available that terminate in \texttt{p(x)} (for True) or \texttt{not p(x)} (for False). If one cannot show \texttt{p(x)} or \texttt{not p(x)} then the answer is Unknown, and again there is not a clear reasoning trace for this; it is necessary to enumerate all possible facts (possibly up to a certain depth) and then if one has not shown  \texttt{p(x)}  or  \texttt{not p(x)} it's considered Unknown. Note that here \texttt{p} is a predicate and \texttt{x} is a variable.

It is for this reason that we used the ProofWriter OWA dataset and removed the Unknowns; this gives us a dataset with reasoning traces concluding in either True or False. If we used CWA we would only have traces that could conclude True.

We evaluate the SI on 5 tasks from the ProofWriter dataset, each requiring varying numbers of reasoning steps (1, 2, 3 and 5). This requires the model to learn to compute intermediate conclusions that may not directly lead to the final answer, but may be needed to reach the final answer. While, this can be very hard to achieve using prompt engineering alone, we endeavour to do so, by demonstrating examples of intermediate and final steps of reasoning (for problems of depth >1). This means that the language model sees (1) examples that both encourage the model to select rules and facts that may not answer the problem in one-step but may help the model to obtain an intermediate output that can be used in a future step and (2) examples of the final step, which takes the model to the final answer. See Sec.~\ref{sec:proof_writer_si_prompt} for an example prompt.

The ProofWriter tasks involve predicting if a given statement, for example \texttt{"Bob is nice."}, is \texttt{True} or \texttt{False} given the context of facts and rules. Our SI model attempts to derive the statement \texttt{"Bob is nice."} or the negation of the statement \texttt{"Bob is not nice."} from the context. To assign a label \texttt{True} or \texttt{False} we follow the procedure proposed in the original ProofWriter paper \citep{tafjord2020proofwriter} and test if any of the implications matches the given statement. If there is a match, the statement is considered to be True, otherwise False. 

ProofWriter results in Fig.~\ref{fig:quant_results} show that the Selection-Inference model outperforms the baselines for problems of depth zero and one, however, with increasing depth, the gap between SI and the baselines diminishes. This is in part because prompt engineering is not sufficient to obtain an optimal Selection module.

Another challenge with the ProofWriter dataset is deciding how many arguments should be selected for each rule. In the ProofWriter dataset, some rules take one argument, others take two. We experimented with various different ways to encourage the model to stop selecting arguments. For example, we append \texttt{". Therefore, "} as a choice to the context that the model can select. If the language model selects \texttt{". Therefore, "} then the selection step ends. We allowed a maximum of two facts to be selected.

\begin{figure}[t!]
    \centering
    \includegraphics[width=0.7\textwidth]{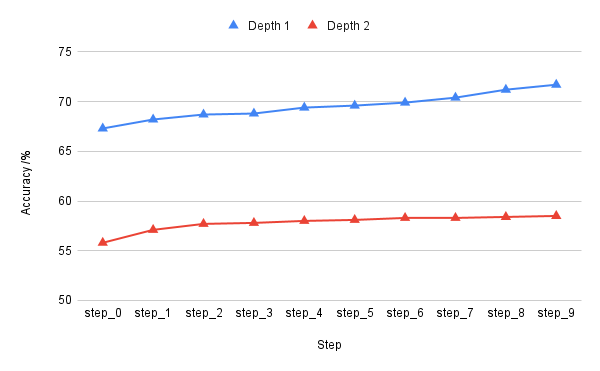}
    \caption{\textbf{ProofWriter: effect of additional reasoning steps.} With additional iterations of selection and inference the probability of the model producing the correct answer increases.}
    \label{fig:effect_of_more_reasoning_steps}
\end{figure}

To obtain results in Fig.~\ref{fig:quant_results} we run SI model for the minimum number of steps needed to solve the problem; a Depth $d$ problem is run for $d$ steps, with the exception of the depth-0 problem which is run for 1 step. However, models may perform better when allowed to run for additional steps, in the case where the model makes a mistake on one step, but later recovers. Fig.~\ref{fig:effect_of_more_reasoning_steps} shows how the number of SI steps can lead to improved performance. There is greater improvement to performance for depth-1 reasoning. For depth-2 reasoning, there was not enough variation in the selections at each step, so additional steps did not help as much as for depth-1.

\section{Reasoning Traces Output by SI}

\subsection{bAbI 15}

Below we show examples of reasoning traces output by our SI framework for the bAbI 15, deductive reasoning task. These examples are ones that the model got correct but are otherwise not cherry picked for their reasoning quality.

\textbf{bAbI 15 Example 1}
\begin{lstlisting}
wolves are afraid of mice
sheep are afraid of mice
winona is a sheep 
mice are afraid of cats
cats are afraid of wolves
jessica is a mouse 
emily is a cat 
gertrude is a wolf 
Question: what is emily afraid of?

Selection: emily is a cat. We know that cats are afraid of wolves
Inference: emily is afraid of wolves
\end{lstlisting}

\textbf{bAbI 15 Example 2}
\begin{lstlisting}
wolves are afraid of mice
sheep are afraid of mice
winona is a sheep 
mice are afraid of cats
cats are afraid of wolves
jessica is a mouse 
emily is a cat 
gertrude is a wolf 
Question: what is winona afraid of?

Selection: winona is a sheep. We know that sheep are afraid of mice
Inference: winona is afraid of mice 
\end{lstlisting}

\textbf{bAbI 15 Example 3}
\begin{lstlisting}
wolves are afraid of mice
sheep are afraid of mice
winona is a sheep 
mice are afraid of cats
cats are afraid of wolves
jessica is a mouse 
emily is a cat 
gertrude is a wolf 
Question: what is gertrude afraid of?

Selection: gertrude is a wolf. We know that wolves are afraid of mice
Inference: gertrude is afraid of mice
\end{lstlisting}

\textbf{bAbI 15 Example 4}
\begin{lstlisting}
wolves are afraid of mice
sheep are afraid of mice
winona is a sheep 
mice are afraid of cats
cats are afraid of wolves
jessica is a mouse 
emily is a cat 
gertrude is a wolf 
Question: what is jessica afraid of?

Selection: jessica is a mouse. We know that mice are afraid of cats
Inference: jessica is afraid of cats
\end{lstlisting}

\subsection{bAbI 16}\label{sec:babi_16_rhino_and_recover_from_error}

In the example below there are two possible conclusions that may be drawn. Since \texttt{brian}, \texttt{bernard} and \texttt{greg} are all rhions but brian and bernard are white, while greg is green the answer could be green or white. A black-box model may select one or the other answer without explaining why. Our model on the other hand shows that it use the facts that \texttt{greg is green} to infer that \texttt{julius is green}.

\textbf{bAbI 16 Example 1: Multiple ``correct'' answers}
\begin{lstlisting}
bernhard is a rhino
brian is a rhino
bernhard is white
brian is white
lily is a lion
lily is yellow
greg is a rhino
greg is green
julius is a rhino
Question: what color is julius?

Selection: julius is a rhino. We know that greg is a rhino and greg is green
Inference: julius is green
\end{lstlisting}

In the example below the model must answer the question \texttt{what color is greg?}. \texttt{greg} is a lion, which means the model needs to infer a rule about lions. However, on the first step of reasoning the model infers that \texttt{we inferred that swans are often gray}. While this is factually correct given the context, it does not help to answer the question. However, the model is able to recover and correctly infer that \texttt{we inferred that lions are often white} meaning that the model is still able to solve the problem despite making a mistake.

\textbf{bAbI 16 Example 2: Recovering from an error}

\begin{lstlisting}
julius is a swan
julius is gray 
bernhard is a swan
bernhard is yellow 
brian is a lion
greg is a lion
brian is white 
lily is a frog
lily is gray 
Question: what color is greg?

Selection: julius is a swan. We know that julius is gray
Inference: we inferred that swans are often gray

Selection: brian is a lion. We know that brian is white
Inference: we inferred that lions are often white
\end{lstlisting}

\section{Fine-tuning Selection-Inference Details and Extra Results}

\begin{figure}[t]
    \centering
         \includegraphics[width=\textwidth]{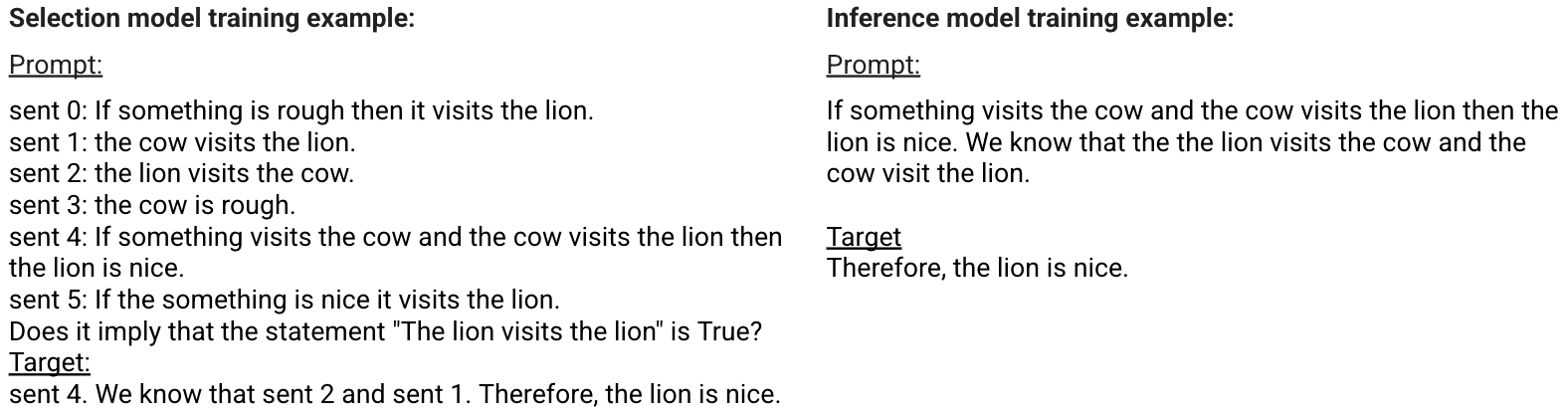}
         \caption{\textbf{Examples of inputs and targets used to fine-tune LLM for the Selection and Inference modules.} On the is an example of supervision for step-1 in a depth-2 problem. We create training data for all steps in the dataset by appending the ground-truth inferences for the intermediate steps to the context. On the right, is an example of a data sample used to train the inference model. Both the Selection and Inference LLM's are fine-tuned on only single steps of reasoning.}
         \label{fig:fine_tune_train_data}
\end{figure}

Fig.~\ref{fig:fine_tune_train_data} shows examples of the format of the data used to fine-tune the Selection and Inference LLMs. The Selection module is trained to predict sentence labels rather than the sentence strings. This prevents the model from making up facts and forces the model to use information in the context.

\begin{figure}[t]

         \centering
         \includegraphics[width=\textwidth]{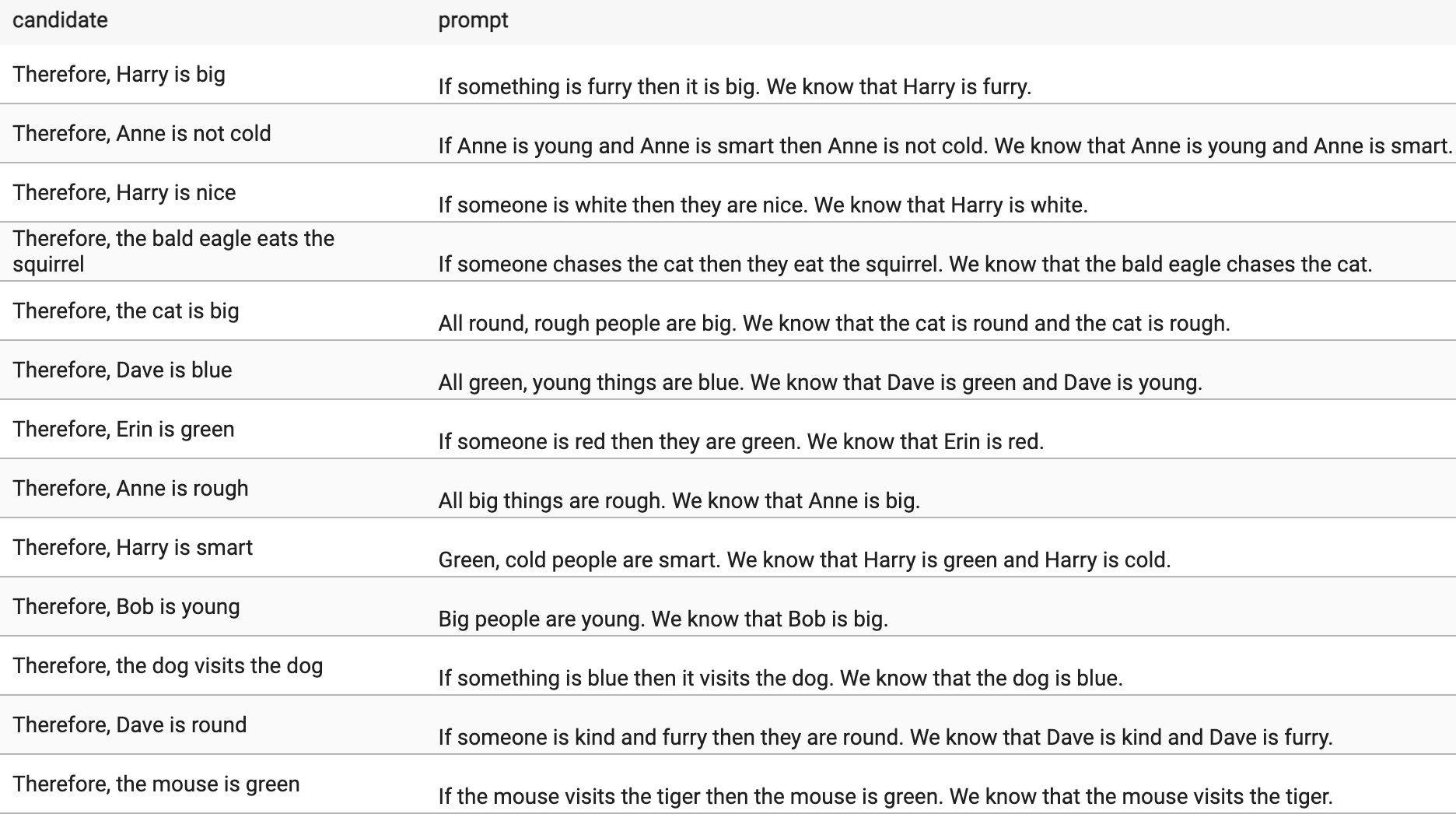}
         \caption{\textbf{Inference Module fine-tuning test examples}}
         \label{fig:fine_tune_inf}
\end{figure}

During selection, rather than scoring each sentence we can append sentence labels to each element in the context (as shown in Fig.~\ref{fig:fine_tune_train_data}); use the Selection LLM to generate selection strings and substitute in elements from the context using a dictionary. This process is much faster than scoring each element of the context, but still ensures that the selection consists only of samples from the context; the Selection module cannot make up facts to answer the question. Fig.~\ref{fig:fine_tune_inf} shows examples of entailment computed by the Inference LLM after fine-tuning.

\begin{figure}[t]
\centering
    \includegraphics[width=\textwidth]{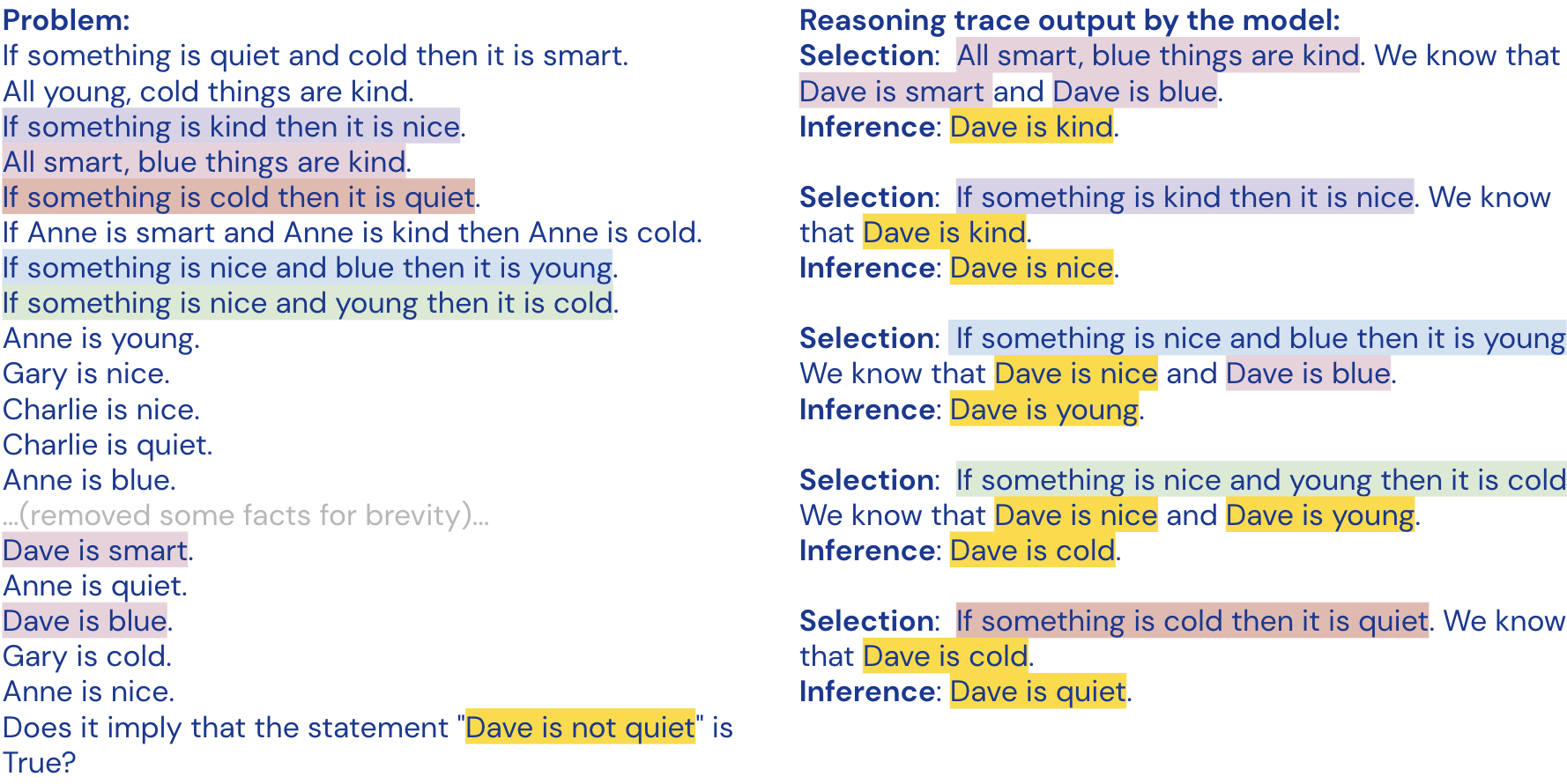}
    \caption{\textbf{A ProofWriter depth-5 reasoning trace output by our model.} The model produces an interpretable reasoning trace that allows us to inspect how the model reached its answer.}   
    \label{fig:fine_tune_depth-5}
\end{figure}

Fig.~\ref{fig:quant_results} compares SI models incorporating fine-tuned vs. prompt-engineered LLMs. We see that the model using LLMs fine-tuned on single steps of reasoning significantly outperform both the prompt-engineered LLMs and a Vanilla LLM prompt-engineered to predict the final answer directly. Fig.~\ref{fig:fine_tune_depth-5} shows a reasoning trace output by the SI model on a challenging depth-5 problem.

\subsection{Selection-Inference Reasoning Traces}\label{sec:si_reason_traces}

\subsubsection{Depth-2 reasoning traces for Depth-2 problems}
Below we show examples of depth-2 reasoning traces produced via Selection-Inference using modules fine-tuned on the ProofWriter dataset.

\textbf{Example 1: }
\begin{lstlisting}
If someone is cold then they eat the lion.
If someone is blue and they chase the dog then the dog chases the lion.
If someone eats the dog then the dog is young.
If someone is young and they eat the lion then they are red.
If someone is nice then they eat the dog.
If someone eats the lion and the lion eats the dog then the dog eats the lion.
If someone sees the lion and the lion chases the dog then the lion is nice.
If the lion sees the dog and the dog sees the lion then the dog is nice.
the lion sees the dog.
the dog sees the lion.
Does it imply that the statement "The dog eats the dog" is True?
step 0:
Selection:  If the lion sees the dog and the dog sees the lion then the dog is nice. We know that the lion sees the dog and the dog sees the lion. 
Inference: The dog is nice.
step 1:
Selection:  If someone is nice then they eat the dog. We know that The dog is nice. 
Inference: The dog eats the dog.
\end{lstlisting}

\textbf{Example 2: }
\begin{lstlisting}
If something is cold and red then it likes the mouse.
If something needs the cat then the cat sees the dog.
If something needs the cow then the cow sees the mouse.
If something sees the dog then the dog likes the cat.
If the cat is not green then the cat does not see the mouse.
If something sees the mouse then it is cold.
If something likes the cat and the cat needs the cow then the cow is nice.
If something sees the cow then it needs the cow.
the mouse needs the dog.
the mouse needs the cat.
the dog is nice.
the cat is green.
the mouse is not nice.
the dog needs the cat.
the dog sees the cow.
the cow is not red.
the cat likes the dog.
the mouse sees the cow.
the mouse needs the cow.
the cow sees the dog.
the mouse is green.
the cow needs the dog.
the mouse is blue.
Does it imply that the statement "The cow is cold" is True?
step 0:
Selection:  If something needs the cow then the cow sees the mouse. We know that the mouse needs the cow. 
Inference: The cow sees the mouse.
step 1:
Selection:  If something sees the mouse then it is cold. We know that The cow sees the mouse. 
Inference: The cow is cold.
\end{lstlisting}

\textbf{Example 3:}
\begin{lstlisting}
Rough things are white.
If Erin is smart and Erin is rough then Erin is white.
If something is round then it is rough.
If Erin is round and Erin is not smart then Erin is white.
All quiet things are not white.
If something is blue and white then it is not quiet.
Erin is green.
Erin is rough.
Erin is blue.
Does it imply that the statement "Erin is not quiet" is True?
step 0:
Selection:  If Erin is smart and Erin is rough then Erin is white. We know that Erin is green and Erin is rough. 
Inference: Erin is white.
step 1:
Selection:  If something is blue and white then it is not quiet. We know that Erin is blue and Erin is white. 
Inference: Erin is not quiet.
\end{lstlisting}

\textbf{Example 4:}
\begin{lstlisting}
If something chases the squirrel then the squirrel is big.
If something is big then it is not kind.
If something chases the bald eagle and it sees the bald eagle then the bald eagle sees the lion.
the cow does not like the squirrel.
the cow sees the lion.
the bald eagle likes the lion.
the cow chases the squirrel.
the lion chases the cow.
the bald eagle is not round.
the squirrel likes the cow.
the cow likes the lion.
the cow chases the bald eagle.
the squirrel likes the bald eagle.
the cow is kind.
the lion chases the squirrel.
the cow does not see the squirrel.
the lion chases the bald eagle.
the squirrel likes the lion.
Does it imply that the statement "The squirrel is kind" is True?
step 0:
Selection:  If something chases the squirrel then the squirrel is big. We know that the cow chases the squirrel. 
Inference: The squirrel is big.
step 1:
Selection:  If something is big then it is not kind. We know that The squirrel is big. 
Inference: The squirrel is not kind.
\end{lstlisting}

\subsubsection{Depth-3 reasoning traces for Depth-3 problems}

Below we show examples of depth-3 reasoning traces produced via Selection-Inference using modules fine-tuned on the ProofWriter dataset.

\textbf{Example 1: }
\begin{lstlisting}
If something eats the bald eagle then it eats the squirrel.
If something eats the squirrel then the squirrel does not eat the bald eagle.
If the bald eagle is not red then the bald eagle likes the squirrel.
If the squirrel likes the bald eagle then the squirrel visits the bald eagle.
If something likes the bald eagle then the bald eagle is young.
If something is young then it eats the squirrel.
the squirrel visits the bald eagle.
the bald eagle visits the squirrel.
the squirrel likes the bald eagle.
Does it imply that the statement "The squirrel does not eat the bald eagle" is True?
step 0:
Selection:  If something likes the bald eagle then the bald eagle is young. We know that the squirrel likes the bald eagle. 
Inference: The bald eagle is young.
step 1:
Selection:  If something is young then it eats the squirrel. We know that The bald eagle is young. 
Inference: The bald eagle eats the squirrel.
step 2:
Selection:  If something eats the squirrel then the squirrel does not eat the bald eagle. We know that The bald eagle eats the squirrel. 
Inference: The squirrel does not eat the bald eagle.
\end{lstlisting}

\textbf{Example 2: }
\begin{lstlisting}
    If someone is quiet then they are white.
If someone is young and red then they are white.
Young people are nice.
If someone is nice then they are round.
All quiet people are young.
Red, big people are nice.
Round, red people are white.
If someone is round then they are quiet.
Harry is young.
Bob is red.
Bob is big.
Does it imply that the statement "Harry is not quiet" is True?
step 0:
Selection:  Young people are nice. We know that Harry is young. 
Inference: Harry is nice.
step 1:
Selection:  If someone is nice then they are round. We know that Harry is nice. 
Inference: Harry is round.
step 2:
Selection:  If someone is round then they are quiet. We know that Harry is round. 
Inference: Harry is quiet.
\end{lstlisting}

\textbf{Example 3:}

Here the model make the same selection on \texttt{step 1} and \texttt{step 2} and therefore does not correctly show that the ``The lion visits the mouse''. Although the model has failed here it is (a) possible that the model can still recover (if ran for additional steps) and compute the correct answer and (b) we can see how the model reached its answer and interpret its failings because the reasoning trace is causal.
\begin{lstlisting}
If something is green then it visits the mouse.
If something chases the bear then it is green.
If something chases the mouse then the mouse sees the bear.
If something sees the bear and the bear visits the mouse then the mouse chases the lion.
If something chases the mouse then it is green.
If something visits the bear then it chases the mouse.
the mouse visits the lion.
the lion visits the bear.
the bear chases the lion.
Does it imply that the statement "The lion does not visit the mouse" is True?
step 0:
Selection:  If something visits the bear then it chases the mouse. We know that the lion visits the bear. 
Inference: The lion chases the mouse.
step 1:
Selection:  If something chases the mouse then it is green. We know that The lion chases the mouse. 
Inference: The lion is green.
step 2:
Selection:  If something chases the mouse then it is green. We know that The lion chases the mouse. 
Inference: The lion is green.
\end{lstlisting}

\section{Limitation Details}\label{sec:limitations}
We have seen that our approach to solving reasoning problems, using SI, has many desirable properties and indeed this model is intended to be a proof of concept to demonstrate that it is possible to build a model with these properties. However, as a proof of concept, this model has several limitations, which we we now discuss in detail.

When observing the outputs of our model, the main point of failure tends to be the Selection module. This is hard to quantify since we do not have labelled data for the intermediate reasoning steps. One reason for this is that we us prompt-engineering to encourage language models to produce the correct outputs, rather than fine-tuning. While our current results are good, and do not require (large amounts of) task specific data, they can be significantly improved by fine-tuning our models for specific tasks, as demonstrated in Section \ref{sec:fine_tuning}.

Prompt engineering works best for single modality cases \citep{nakano2021webgpt}. It is more difficult to get the model to do multi-step reasoning since the distribution or patterns for the intermediate steps differ to the final step. It is also difficult to get the model to figure out how many arguments to select or how many arguments a rule takes, using only prompt engineering, again because there are multiple different patterns that the model needs to learn how and when to apply.

Other limitations of our work include the assumption that a database of facts or rules are given. In many practical settings we would need to be able to retrieve relevant knowledge from an existing knowledge base. There is exciting progress being made in this area \citep{lazaridou2022internet} and we hope in the future to combine these approaches with the SI model.

Finally, while our model has the benefit that performance scales with compute time; the longer we run our model the more likely it is to reach a correct answer, we do not have a good way of deciding when to halt the reasoning process or to filter the reasoning traces. In our current approach we have fixed budget, and tend to report results based on the final inference. Results in Fig.~\ref{fig:effect_of_more_reasoning_steps} suggests that accuracy could be improved if we had a mechanism for filtering the reasoning steps and selecting the best answer.

\section{Baseline Datasets}
\label{si_sec:baseline_results}
In this paper we used tasks from six sources: bAbi \citep{weston2015towards}, BigBench \citep{BigBench}, AAC \citep{betz2020critical}, Jeopardy \citep{tunguz_2019}, ProofWriter \citep{tafjord2020proofwriter} and 2WikiMultiHop \citep{welbl2018constructing}. All of these dataset are publicly available and our use of the data was in accordance to their respective license permissions. As far as we are aware, none of the datasets contain personally identifiable information or offensive content.

\paragraph{Task decomposition} From bAbI dataset \citep{weston2015towards} used 
tasks 1-3 to measure the ability of LLMs to cope with progressively larger numbers of reasoning steps on the same type of problem; 
task 6 to compare to task 1 whether yes/no questions are easier for the models to answer compared to more free-form answers; 
task 9 to compare to task 1 to test how well the models can deal with negation; 
task 10 to test whether LLMs know that the facts they are given are not sufficient to answer a question;
tasks 15 and 16 to test basic deduction and induction abilities respectively;
task 18 to compare to task 2 for two step reasoning on a different kind of task (based on size).

Jeopardy \citep{tunguz_2019} and 2WikiMultiHop \citep{welbl2018constructing} tasks measure the ability of LLMs to do reasoning in less structured settings, where the answer has to be generated in free form, and the level of available context varies between none (2WikiMultiHop, Jeopardy), to relevant and irrelevant unstructured context (2WikiMultiHop With Context and 2WikiMultiHop With Context \& Facts), to relevant structured context only (2WikiMultiHop With Evidence).

AAC \citep{betz2020critical} measures the ability of LLMs to do relatively shallow formal reasoning (1-2 steps) over a relatively large set of syllogistic argument schemes, both with (AAC Split Extended) and without (AAC Split) dealing with negation.

ProofWriter \citep{tafjord2020proofwriter} tasks measure the ability of models to do formal reasoning over a progressively more difficult tasks that require more steps of reasoning.

From BigBench \citep{BigBench} we imported the following tasks:
Analytic Entailment, Epistemic Reasoning and  measure the ability of LLMs to decide implicit entailment relationship given facts.

Entailed Polarity, Presuppositions as NLI and Logical Arguments measure the ability of LLMs to understand implied information from vague language.

Evaluating Information Essentiality and Sufficient Information measure how well LLMs can evaluate which context information is relevant and sufficient to answer a question.

Formal Fallacies Syllogisms Negation, Logic Grid Puzzle, Logical Fallacy Detection, and Logical Deduction test the ability of LLMs to do formal deductive reasoning. 

Sequence Problems Tasks and Tracking Shuffled Objects are similar to bAbI tasks 1-3 and evaluates the ability of LLMs to do multi-hop reasoning based on sequenced facts.

Physics Questions and Unit Interpretation measure the ability of LLMs to reason about grade school science problems.

StrategyQA measures the ability of LLMs to do multi-hop reasoning based on general knowledge that is not explicitly provided as context.

\paragraph{Multiple choice normalisation} We evaluated whether normalising log probability of the choices under the model by the token length resulted in better accuracy to avoid potential bias as reported in \citep{lin2020birds}. We found significant ($p=0.0002$, two-tailed t-test with equal variance) but minimal difference between the average accuracy across all evaluated tasks, when evaluated with ($67.92\pm 46.68 \%$) and without normalisation ($68.3\pm 46.53 \%$). For this reason we use the unnormalised measures in the paper.

\paragraph{Evaluating dataset bias} To check whether the datasets we use are biased, we compared how much the baseline multiple choice accuracy of LLMs when presented with options all by themselves, without any context or question, deviates from the expected random performance calculated as $1/N$, where $N$ is the number of choices. We found that the two differed by a very small amount $0.08 \pm 8.74 \%$ on average across all models and all multiple-choice datasets ($p=5e-16$, two-tailed t-test with equal variance). Since the effect size was so small, we concluded that the datasets were not biased and present the expected random baseline where appropriate. 

\paragraph{Appending choices to bAbI tasks} We evaluated whether adding choices to the prompt improved the multiple choice accuracy of LLMs on bAbI tasks. We found that this was not the case, with the average accuracy across all bAbI tasks being $37.86\pm 48.5 \%$ when choices were appended, and $44.86 \pm 49.74 \%$ when choices were not appended ($p=2e-61$, two-tailed t-test with equal variance). For this reason, we report the latter results in the paper.

\paragraph{2WikiMultiHop results} We evaluated the performance of LLMs on 2WikiMultiHop \citep{welbl2018constructing} dev subset using exact string match between the generated and the ground truth answers. In particular, the generated answer was truncated at the first sentence up to ".", "?", "!", ";" or newline characters following the BigBench generative evaluation protocol \citep{BigBench}. The two answers were then both stripped of all punctuation, white space and special characters before comparison is made. Dataset examples receive a score of 1 if the post-processed answers match exactly, and 0 otherwise. 

We found that the models scored $13.62\pm34.3 \%$ on average on the original dataset, consisting of questions only. When the context of Wikipedia paragraphs with relevant and irrelevant facts to answer the question was added to the context, the performance dropped to $1.47\pm12.02 \%$ on average. Adding information about the relevant facts within these context paragraphs did not help much, resulting in $1.97\pm13.9 \%$ accuracy. On the other hand, adding only the relevant facts extracted from the underlying knowledge graph triples has more than doubled the models' performance, resulting in $35.55\pm 47.87 \%$ average accuracy.

\paragraph{General insights}
LMs get progressively worse as more steps are needed for reasoning (see bAbI tasks 1-3 and 18, ProofWriter tasks; although not the case for Logical Deduction, Sequence Problems Tasks and Tracking Shuffled Objects is at chance).

LMs find yes/no questions harder to answer than freeform questions (see bAbI task 6 vs 1).

LMs struggle to deal with negation (bAbi task 9 vs 1) unless it is in a very well formalised limited setup (AAC Split Extended).

LMs struggle with deciding when they do not have sufficient information (bAbI task 10 vs 1; Evaluating Information Essentiality and Sufficient Information are both at chance).

LMs are average at formal deduction and induction tasks, although their performance very much depends on the difficulty of the task and the evaluation protocol (see bAbI task 15 and Logical Deduction, although Proof Writer, Formal Fallacies Syllogisms Negation, Logic Grid Puzzle, and Logical Fallacy Detection results are close to chance, while AAC results, where the models are evaluated in a very structured setting are very good).

In less formal mutli-hop question answering scenarios, LLMs are close to chance when no context is provided (2WikiMultiHop, StrategyQA; although Jeopardy is an outlier) or when the provided information is unstructured (e.g. whole Wikipedia paragraphs as in 2WikiMultiHop With Facts and 2WikiMultiHop With Facts \& Rules), but do better when minimal structured context information is provided (2WikiMultiHop With Evidence). 

LMs also perform poorly in solving grade school science problems (Physics Questions and Unit Interpretation) although this ability is better in multiple choice compared to generative evaluation settings.

LMs are, however, reasonable at doing simple implication, entailment and induction tasks (see bAbI task 16, Analytic Entailment, Entailed Polarity, and Logical Arguments; although on Epistemic Reasoning and Presuppositions as NLI the models perform around chance level).

\begin{figure}
    \begin{subfigure}[b]{0.49\textwidth}
    \centering
    \includegraphics[width=0.99\textwidth]{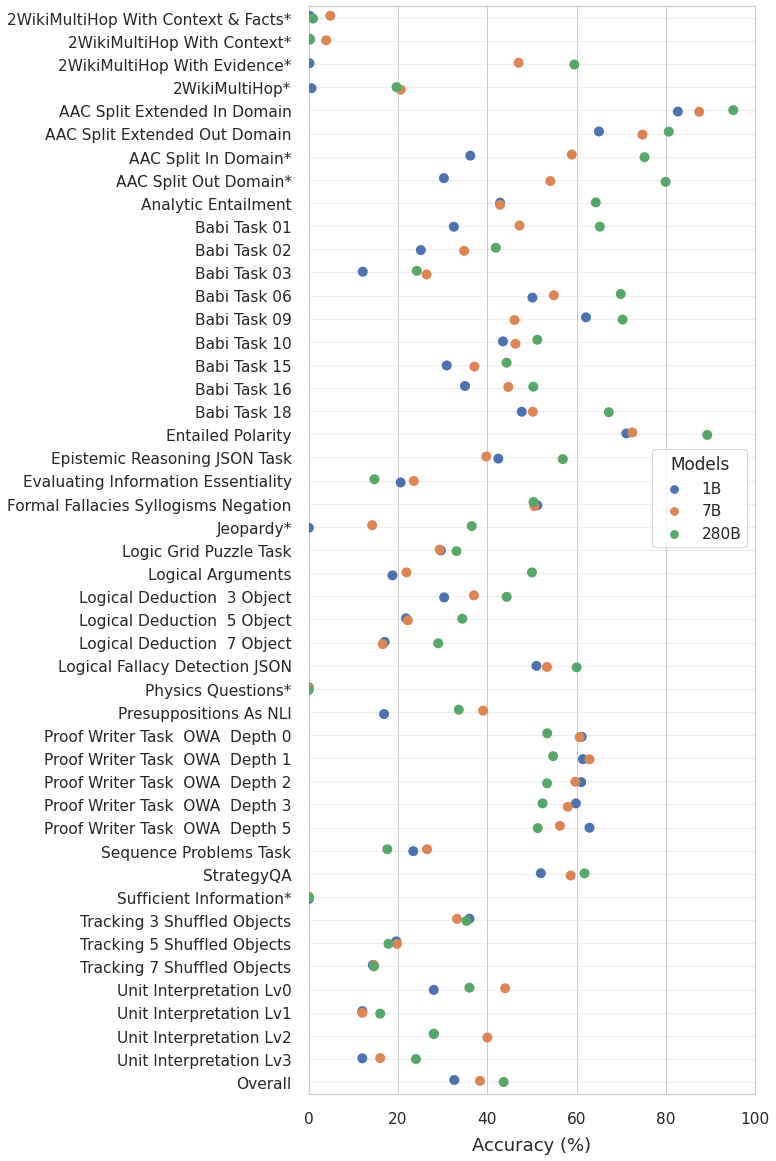}
    \caption{Absolute multi-choice and generative (marked with an asterix *) accuracy.}
    \label{fig:logic_baselines}
    \end{subfigure}
    \hspace{0.5cm}  
    \begin{subfigure}[b]{0.49\textwidth}
    \centering
    \includegraphics[width=0.99\textwidth]{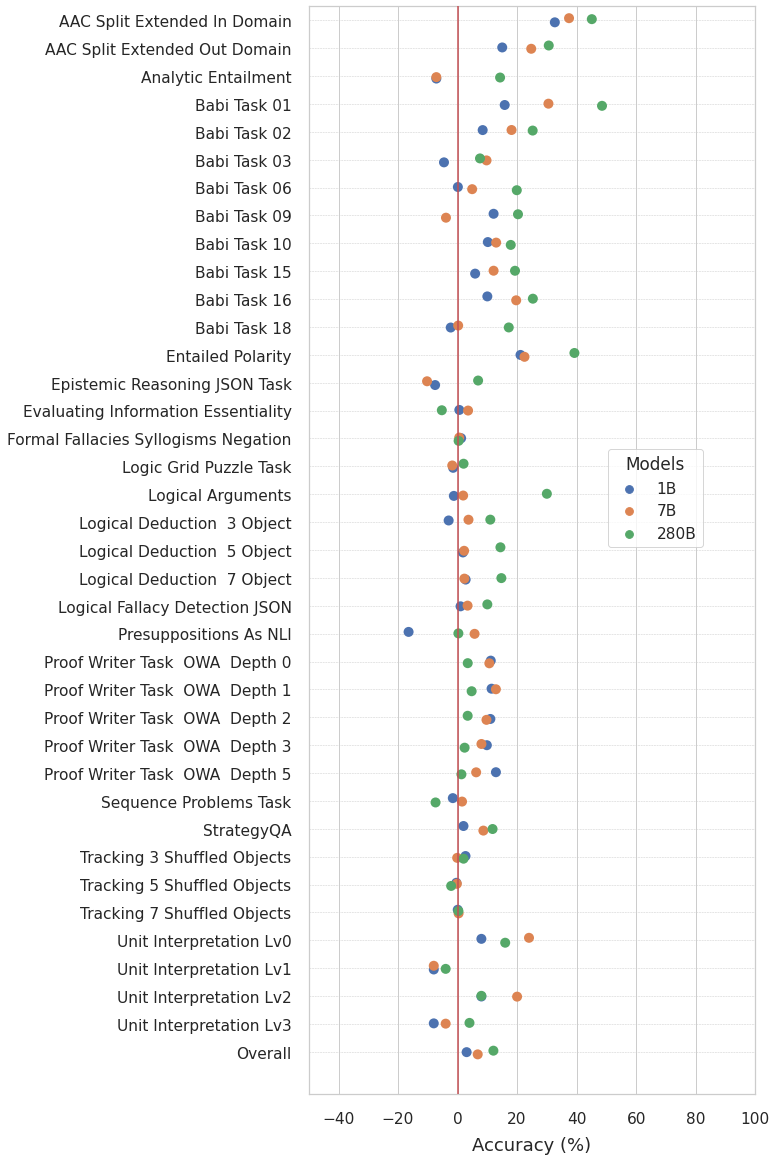}
    \caption{Relative accuracy compared to chance level for multi-choice tasks. Red line - chance performance.}
    \label{fig:logic_baselines_multichoice_less_random}
    \end{subfigure}
    \caption{Average accuracy of LLMs from the Gopher family \citep{rae2021scaling} evaluated on logical reasoning tasks in a 5-shot generalisation setting.}
\end{figure}

\section{Tests of statistical significance}
To calculate statistical significance of differences between different models in Fig.~\ref{fig:si_results} we used proportion hypothesis test for binary data. In particular, we used two-sided \texttt{proportions\_ztest} from \texttt{statsmodels.stats.proportion}.

\end{document}